\documentclass[preprint,12pt]{elsarticle}
\usepackage{amssymb}
\usepackage{amsmath}
\usepackage{hyperref}
\usepackage{array}
\usepackage{mdwmath}
\usepackage{mdwtab}
\usepackage{eqparbox}
\usepackage{url}
\usepackage{float}
\usepackage{algorithm}
\usepackage{algorithmic}
\usepackage{lipsum} 

\usepackage{booktabs} % For formal tables
\usepackage{tabularx}
\usepackage{caption}
\usepackage{multirow}
\usepackage{hyperref}
\usepackage{pifont}
\usepackage{fancyhdr}

\usepackage[normalem]{ulem}
\usepackage{xcolor}

\usepackage{soul}   
\usepackage{etoolbox}

\begin{document}

\begin{frontmatter}

\title{ABG-NAS: Adaptive Bayesian Genetic Neural Architecture Search for Graph Representation Learning}

\author[ltu]{Sixuan Wang}
\author[vu]{Jiao Yin}
\author[ltu]{Jinli Cao\corref{cor1}}
\author[ltu]{Mingjian Tang}
\author[vu]{Hua Wang}
\author[zjnu]{Yanchun Zhang}

\cortext[cor1]{Corresponding author: j.cao@latrobe.edu.au}

\affiliation[ltu]{
  organization={Department of CSIT, La Trobe University},
  city={Melbourne},
  state={VIC},
  postcode={3086},
  country={Australia}
}

\affiliation[vu]{
  organization={Institute for Sustainable Industries and Liveable Cities (ISILC), Victoria University},
  city={Melbourne},
  state={VIC},
  postcode={3011},
  country={Australia}
}

\affiliation[zjnu]{
  organization={School of Computer Science and Technology, Zhejiang Normal University},
  city={Jinhua},
  state={Zhejiang},
  postcode={321004},
  country={China}
}

\begin{abstract}

Effective and efficient graph representation learning is essential for enabling critical downstream tasks, such as node classification, link prediction, and subgraph search. However, existing graph neural network (GNN) architectures often struggle to adapt to diverse and complex graph structures, limiting their ability to produce structure-aware and task-discriminative representations. To address this challenge, we propose ABG-NAS, a novel framework for automated graph neural network architecture search tailored for efficient graph representation learning. ABG-NAS encompasses three key components: a Comprehensive Architecture Search Space (CASS), an Adaptive Genetic Optimization Strategy (AGOS), and a Bayesian-Guided Tuning Module (BGTM). CASS systematically explores diverse propagation (\textbf{P}) and transformation (\textbf{T}) operations, enabling the discovery of GNN architectures capable of capturing intricate graph characteristics. AGOS dynamically balances exploration and exploitation, ensuring search efficiency and preserving solution diversity. BGTM further optimizes hyperparameters periodically, enhancing the robustness and scalability of the resulting architectures to both large-scale graphs and high-complexity models. Empirical evaluations on benchmark datasets (Cora, PubMed, Citeseer, and CoraFull) demonstrate that ABG-NAS consistently outperforms both manually designed GNNs and state-of-the-art neural architecture search (NAS) methods. These results highlight the potential of ABG-NAS to advance graph representation learning by providing adaptive solutions that scale effectively across varying graph sizes and architectural complexities. Our code is publicly available at \href{https://github.com/sserranw/ABG-NAS}{https://github.com/sserranw/ABG-NAS}.

\end{abstract}

\begin{keyword}
Graph representation learning \sep neural architecture search \sep graph neural networks \sep bayesian optimization \sep genetic algorithms
\end{keyword}

\end{frontmatter}

\section{Introduction}
Graph representation learning has become essential for enabling critical tasks, such as node classification, graph classification, and subgraph search, by leveraging graph-structured data \cite{zhili2024search,huang2025transformer}. Graph neural networks (GNNs) are at the forefront of this field due to their ability to aggregate and transform node features based on graph topology \cite{yin2024heterogeneous,hong2022graph,khan2024synergies}. However, the performance of GNNs is heavily influenced by their architectural design. Traditional GNNs with fixed propagation (\textbf{P}) and transformation (\textbf{T}) components often lack the flexibility needed to adapt to diverse and complex real-world graphs, which exhibit varying node degree distributions and connectivity patterns \cite{qin2022bench}. This rigidity limits their generalizability across different tasks, and manually designing architectures for specific graphs is both time-intensive and demands substantial expertise \cite{peng2024graphrare}. These challenges are further compounded by the inherent diversity of graph structures, ranging from sparse graphs with low connectivity to dense graphs with highly interconnected nodes, which demand architectures capable of adapting to both extremes\cite{chen2024evoprompting}.

Graph neural architecture search (GNAS) has recently emerged as a powerful paradigm that eliminates manual trial-and-error by automatically searching for GNN architectures adaptable to a wide range of structural complexities and representation tasks \cite{chen2024evoprompting}. It significantly reduces the reliance on manual design and enables more scalable solutions in graph representation learning. Existing GNAS methods can be broadly categorized into three types \cite{oloulade2021graph}: reinforcement learning (RL)-based methods \cite{fu2022reinforced}, which optimize the search process through policy optimization; gradient-based methods, which leverage differentiable architecture search techniques; and evolutionary algorithm (EA)-based methods, including genetic algorithms (GA), particle swarm optimization (PSO), and differential evolution (DE).

RL-based methods have shown promise in GNAS by automating the search process through systematic exploration of the architecture space. Typically, GraphNAS\cite{gao2021graph} employs a recurrent neural network (RNN) to sequentially construct GNN architectures, enabling efficient exploration of the search space and providing flexible solutions for diverse datasets. Subsequently, simplified neural architecture search for graph neural networks (SNAG) \cite{zhao2020simplifying} extended this line of research by introducing macro-level inter-layer architecture connections, offering a more comprehensive exploration of GNN designs. However, RL-based approaches are often constrained by high computational demands and fixed search spaces\cite{chen2023mngnas}, which limit the diversity of generated architectures and reduce their adaptability to complex and diverse graph structures \cite{yin2024compact,li2023survey,hong2023graph}.

In response to these constraints, gradient-based NAS methods such as search to aggregate NEighborhood (SANE) \cite{huan2021search} and automated graph neural network on heterophilic graphs (Auto-HeG) \cite{zheng2023auto} have been developed. These methods leverage differentiable search spaces, allowing for continuous optimization via gradient descent techniques. Although gradient-based approaches improve search efficiency and can discover diverse architectures, they still struggle with issues like local optima and high computational costs, which can limit their practical application.

As a result, EA-based NAS methods have gained increasing attention \cite{zhang2022dfg,shi2022genetic}. EA-based approaches facilitate a more comprehensive exploration of the search space, increasing the likelihood of finding global optima\cite{feng2024enhancing}. By allowing the combination of various \textbf{P} and \textbf{T} operations tailored to different graphs, EA-based methods provide a broader and more flexible approach to GNAS. For example, Zhang et al. \cite{zhang2022dfg} introduced a method that flexibly combines single-type propagation and transformation components to construct adaptive architectures. However, due to the simplicity of their search spaces and strategies, these approaches have not consistently outperformed manually designed architectures.

Despite the strengths of EA-based approaches in node classification tasks, they still face significant challenges. Real-world graphs with varying sparsity and density present unique challenges for graph data management, requiring architectures that can generalize across diverse tasks while maintaining efficiency and accuracy. However, many current GNAS methods rely on limited search spaces and static search strategies \cite{zhang2022dfg}, which hinder their adaptability to such diverse and complex graph structures. Furthermore, the critical aspect of hyperparameter optimization is often neglected \cite{shi2022genetic}, leading to suboptimal performance across different graph types.

To address these challenges, we propose ABG-NAS, a novel adaptive Bayesian genetic (ABG) algorithm-based GNAS framework tailored for efficient graph representation learning. ABG-NAS integrates a Comprehensive Architecture Search Space (CASS), an Adaptive Genetic Optimization Strategy (AGOS), and a Bayesian-Guided Tuning Module (BGTM). The main contributions of the proposed ABG-NAS framework are threefold:

\begin{itemize}
    \item \textbf{Comprehensive Architecture Search Space:}  We generalize the decomposition of propagation (\textbf{P}) and transformation (\textbf{T}) operations from GCNs to broader GNN architectures and expand the search space to include various internal configurations of \textbf{P} and \textbf{T} within each layer. This approach enables the exploration of both diverse layer sequences and distinct internal setups of \textbf{P} and \textbf{T}. This comprehensive expansion of the search space enables a more thorough exploration of GNN architectures, enhancing their adaptability to diverse real-world scenarios.

    \item \textbf{Adaptive Genetic Optimization Strategy:} Unlike traditional genetic algorithms with static crossover and mutation rates, AGOS intelligently re-calibrates these rates, dynamically optimizing the balance between exploration and exploitation. These designs significantly reduce redundant evaluations, ensuring both population diversity and improved search efficiency.
    
    \item \textbf{Bayesian-Guided Tuning Module:} We seemlessly embed a Bayesian optimization strategy within the AGOS search process at periodic intervals to selectively optimize hyperparameters of the identified GNN architectures and training processes, balancing efficiency and performance. 
   
\end{itemize}

To validate the effectiveness of the proposed ABG-NAS framework, we conduct comprehensive experiments and ablation studies on four benchmark datasets: Cora, PubMed, Citeseer, and CoraFull. The results demonstrate that ABG-NAS consistently outperforms manually designed GNN architectures and existing automated GNAS methods in graph representation learning tasks. This highlights its capability to provide adaptable and high-performing solutions across diverse graph datasets.

The paper is organized as follows. Section \ref{sec:RL} reviews the background, related works, and preliminaries. Section \ref{sec:proposed-algorithm} describes the ABG-NAS framework in detail, including the implementation of its main components. Section \ref{sec:exp} presents and analyzes the experimental results from multiple perspectives. Finally, Section \ref{sec:conclusion} summarizes the findings of this paper and discusses future work.

\section{Related Works}
\label{sec:RL}
% \subsection{Manual Design of GNN Architectures}
\subsection{GNNs in Graph Representation Learning}

\begin{figure}[ht]
\centering
\includegraphics[width=\textwidth,clip,trim=5 5 5 5]{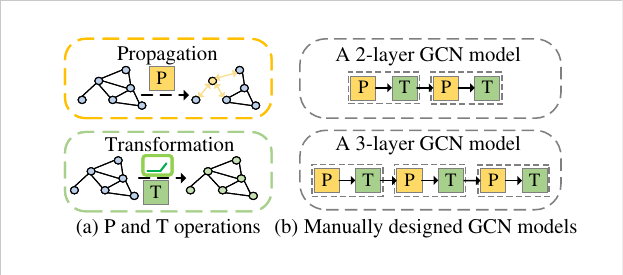}  
\caption{The illustration of \textbf{P} and \textbf{T} operations in a GCN layer and their use in manually-designed models.}
\label{fig:PT-process}
\end{figure}

Graph representation learning transforms graph-structured data into meaningful representations, forming the foundation for advanced methods such as graph neural networks (GNNs)\cite{zhili2024search,hong2022graph}. GNNs, as one of the most prominent approaches, have gained attention for their ability to model intricate relationships in graphs through learnable layers\cite{wang2023bayesian,you2023knowledge}. Researchers have devised various GNN models with different types of layers, such as graph convolutional network (GCN) \cite{kipf2016semi}, graph attention network (GAT) \cite{velickovic2017graph},  graph sample and aggregation (GraphSAGE) \cite{hamilton2017inductive}, and modality-aware graph convolutional network (MAGCN) \cite{yin2022knowledge}, allowing for manually flexible stacking of different layers tailored to specific applications. Furthermore, Zhang et al. (2022) \cite{zhang2022dfg} were the first to try to decompose a GCN layer into two distinct operations: \textbf{P} and \textbf{T}. 
As demonstrated in Figure \ref{fig:PT-process} (a), the \textbf{P} operation aggregates the features from  neighouring nodes to update the representation of nodes. Following this, the \textbf{T} operation applies a learnable weight matrix and a nonlinear activation function to enhance feature representation, allowing the GCN to capture more complex patterns.

Specifically, for a given graph \(\mathcal{G} = (V, E, \mathbf{X})\), where \(v_i \in V, \ i = 1, \ldots, N\) represents the \(i\)-th node and \(N\) is the total number of nodes, \(E \subseteq V \times V\) represents the set of edges, \(\mathbf{X} = \{\mathbf{x}_1, \mathbf{x}_2, \ldots, \mathbf{x}_N\}\) is the feature matrix of V, and \(\mathbf{A} \in \mathbb{R}^{N \times N}\) is the adjacency matrix of the graph. The adjacency matrix \(\mathbf{A}\) reflects the edge relationships between nodes, where \(\mathbf{A}_{ij} \neq 0\) if there is an edge \( e_{ij} \in E\) between nodes $v_i$ and $v_j$, and \(\mathbf{A}_{ij} = 0\) otherwise. The\textbf{ P} and \textbf{T} operations of a GCN layer can be presented as Equations (\ref{eq:propagation}) and (\ref{eq:transformation}), respectively.

\begin{equation}
m^l_v = \psi \left( \{ h^{l-1}_u \mid u \in N_v \} \right)
\label{eq:propagation}
\end{equation}
\begin{equation}
h^l_v = \delta \left( W^l m^l_v \right)
\label{eq:transformation}
\end{equation}

In Equation (\ref{eq:propagation}), the \textbf{P} operation aggregates the features \( h^{l-1}_u \) from neighboring nodes \( u \in N_v \) to update node \( v \)'s representation \( m^l_v \), where \(\psi(\cdot)\) is the aggregation function, which typically involves operations like summing, averaging, or taking the maximum of the neighboring features. The\textbf{ T} operation, represented in Equation (\ref{eq:transformation}), applies a learnable weight matrix \( W^l \) and a nonlinear activation function \( \delta(\cdot) \) to transform the intermediate representation \( m^l_v \) into the final node feature representation \( h^l_v \). In this context, \(l\) denotes the layer index, indicating the specific layer of a GCN model.

Previously, GCN architectures were limited to a rigid (\textbf{P-T}) sequence pattern, such as (\textbf{P-T})\textbf{-}(\textbf{P-T}) for a 2-layer GCN and \textbf{(P-T})\textbf{-}(\textbf{P-T})\textbf{-}(\textbf{P-T}) for a 3-layer GCN, as shown in Figure \ref{fig:PT-process} (b). By decomposing layers into separate P and T operations, a wider variety of configurations, such as \textbf{P-P-T-T}, \textbf{P-P-T}, and \textbf{T-T-P-T-P}, can be constructed, allowing for more modular and adaptable GNN architectures.

While this decomposition enhances flexibility, it also complicates manual design, making it challenging to identify optimal configurations. Automated NAS algorithms are therefore needed to explore this expanded architectural space and tailor GNN structures to diverse tasks and datasets. This motivates the adoption of EA-based GNAS, which can efficiently search and optimize GNN designs across a broad range of possibilities.

\subsection{EA-based GNAS}
\label{subsec:EA-background}

Among evolutionary algorithms (EAs), genetic algorithm (GA) has proven particularly effective in addressing the combinatorial optimization challenges inherent in graph neural architecture search (GNAS)\cite{oloulade2021graph}. For example, the deep and flexible graph neural architecture search (DFG-NAS) framework \cite{zhang2022dfg} successfully applied GA to optimize P and T operations in GNAS, showcasing its capability to solve complex search and optimization problems. To comprehensively explain the GA-based GNN automation process in DFG-NAS\cite{zhang2022dfg}, we summarize its six-step implementation. 

The process begins with the initialization of the population, denoted as \( \mathcal{Q}_{p}^0 = \{ I_{pi}^0\}_{i=1}^{N_s} \). \( \mathcal{Q}_{p}^0 \) contains \( N_s \) individuals. Each individual \( I_{pi}^0 \) is represented as a sequence randomly assembled from the operation set \( \mathcal{O} = \{ \textbf{P}, \textbf{T} \} \), with its length also randomly determined. The subscript \( p \) denotes the parent population. Next, the fitness evaluation begins. For each individual in the \( g \)-th generation (\(0 \leq g \leq N_g \)), the fitness is calculated by decoding the individual into a GNN architecture, training it on a node classification task, and evaluating its test accuracy. The fitness scores for the population are denoted as \( \mathcal{S}_{p}^g = \{ s_{pi}^g\}_{i=1}^{N_s} \), where each \( s_{pi}^g \) represents the fitness score of the \( i \)-th individual. Once fitness is evaluated, the process moves to offspring generation. DFG-NAS employs a 3-tournament selection strategy. Specifically, three individuals are randomly selected from the population \( \mathcal{Q}_{p}^0 \), and the one with the highest fitness (i.e., accuracy) is chosen as the parent for evolution, denoted as \( I_{p}^g\). After selecting the parent, mutation is applied to create a new offspring. Mutation involves three equally probable operations: adding, removing, or swapping \textbf{P} and \textbf{T} components within this encoded GNN architecture, generating an offspring \( I_{o}^g \). This offspring then is decoded into a GNN architecture and its fitness value \( s_{o}^g \) is calculated using the same procedure as the parent. To maintain a constant population size, environmental selection is performed. The newly generated offspring \( I_{o}^g \) is added to the current parent population \( \mathcal{Q}_{p}^g \), and the oldest individual \( I_{p_1}^g \) is removed. This process results in the next generation's parent population, denoted as \( \mathcal{Q}_{p}^{g+1} = (\mathcal{Q}_{p}^{g} \cup \{ I_{o}^g \}) \setminus \{ I_{p_1}^g \} \). This process iterates until \( g == N_g \), at which point the algorithm outputs \( I_{o}^g \) as the optimal architecture for automating the design of GNN structures. Despite its pioneering approach, DFG-NAS has several limitations, including a simplistic search space, basic GA operations, and the lack of hyperparameter optimization during GCN training, presenting opportunities for further research and enhancement.

\subsection{Hyperparameter Optimization for GNAS}
\label{subsec:HPO-background}

Hyperparameter Optimization (HPO) is an optional component of GNAS algorithms, used to optimize the training hyperparameters of a given architecture, aiming to identify an effective combination of hyperparameters. Shi et al. \cite{shi2022genetic} successfully applied GA to simultaneously optimize both the architecture and its training hyperparameters within NAS tasks. This integrated approach enhances the compatibility between the architecture and its hyperparameters, leading to improvements in model convergence speed and overall performance. However, the GA-based approach, which alternates between optimizing architectures and hyperparameters independently, can suffer from slower convergence and higher computational costs, especially in large search spaces.

In hyperparameter optimization tasks, Bayesian Optimization (BO) has been widely recognized as an efficient optimization method\cite{onorato2024bayesian}. BO combines surrogate models and \textbf{acquisition functions} to search for optimal solutions effectively in the hyperparameter space. For the surrogate model, BO often employs Gaussian Processes (GPs) or Tree-structured Parzen Estimators (TPE). TPE models the conditional probability distribution \( p(\mathbf{x} | y) \), defined as Equation\ref{eq:TPE}:

\begin{equation}
p(\mathbf{x} | y) =
\begin{cases}
l(\mathbf{x}) & \text{if } y < y^* \\
g(\mathbf{x}) & \text{if } y \geq y^*
\end{cases}
\label{eq:TPE}
\end{equation}

Here, \( y^* \) represents the performance threshold, while \( l(\mathbf{x}) \) and \( g(\mathbf{x}) \) denote the distributions of configurations below and above this threshold, respectively.

In addition, BO leverages acquisition functions, such as Expected Improvement (EI), to determine the next promising hyperparameter configuration, defined as \( EI(\mathbf{x}) = \mathbb{E}[\max(0, y^* - f(\mathbf{x}))] \), where \( y^* \) represents the current best objective value, and \( f(\mathbf{x}) \) is the predicted objective value for the hyperparameter configuration \( \mathbf{x} \).

Recent research has extensively explored ways to improve BO's performance, focusing on enhancing surrogate models or refining the design of acquisition functions. For instance, PFNs4BO\cite{muller2023pfns4bo} and HyperBO\cite{wang2024pre} incorporate pretrained models and data-driven GP priors, significantly improving BO's efficiency in high-dimensional problems and modeling complex prior distributions. However, these methods primarily target static optimization tasks and remain less adaptable to dynamic resource-constrained or real-time scenarios. Furthermore, FanG-HPO\cite{candelieri2024fair} extends BO to support multi-objective optimization (e.g., accuracy, fairness, and resource efficiency), providing flexible solutions for resource-constrained practical scenarios. However, the design of FanG-HPO does not fully consider the balance between exploration and exploitation at different optimization stages, potentially limiting its efficiency in complex tasks. FT-PFN~\cite{rakotoarison2024context} addresses this by leveraging its in-context learning capability to avoid the online training overhead of Gaussian surrogates, and improving the acquisition function to enhance the balance between exploration and exploitation, thereby significantly boosting hyperparameter optimization efficiency in resource-constrained scenarios. Nevertheless, FT-PFN's phased strategy mainly applies to fixed architecture optimization tasks and has not yet been extended to the joint optimization of architectures and hyperparameters.

Although BO-based HPO approaches have achieved certain advancements, no existing work has attempted to extend HPO to the joint optimization of architectures and hyperparameters. Future research should focus on developing efficient strategies to seamlessly integrate BO-HPO processes into GA-based GNAS frameworks, thereby optimizing computational efficiency.

\section{Proposed Method}
\label{sec:proposed-algorithm}

\begin{figure}[H]
    \centering
    \includegraphics[width=\textwidth,clip,trim=2 2 2 2]{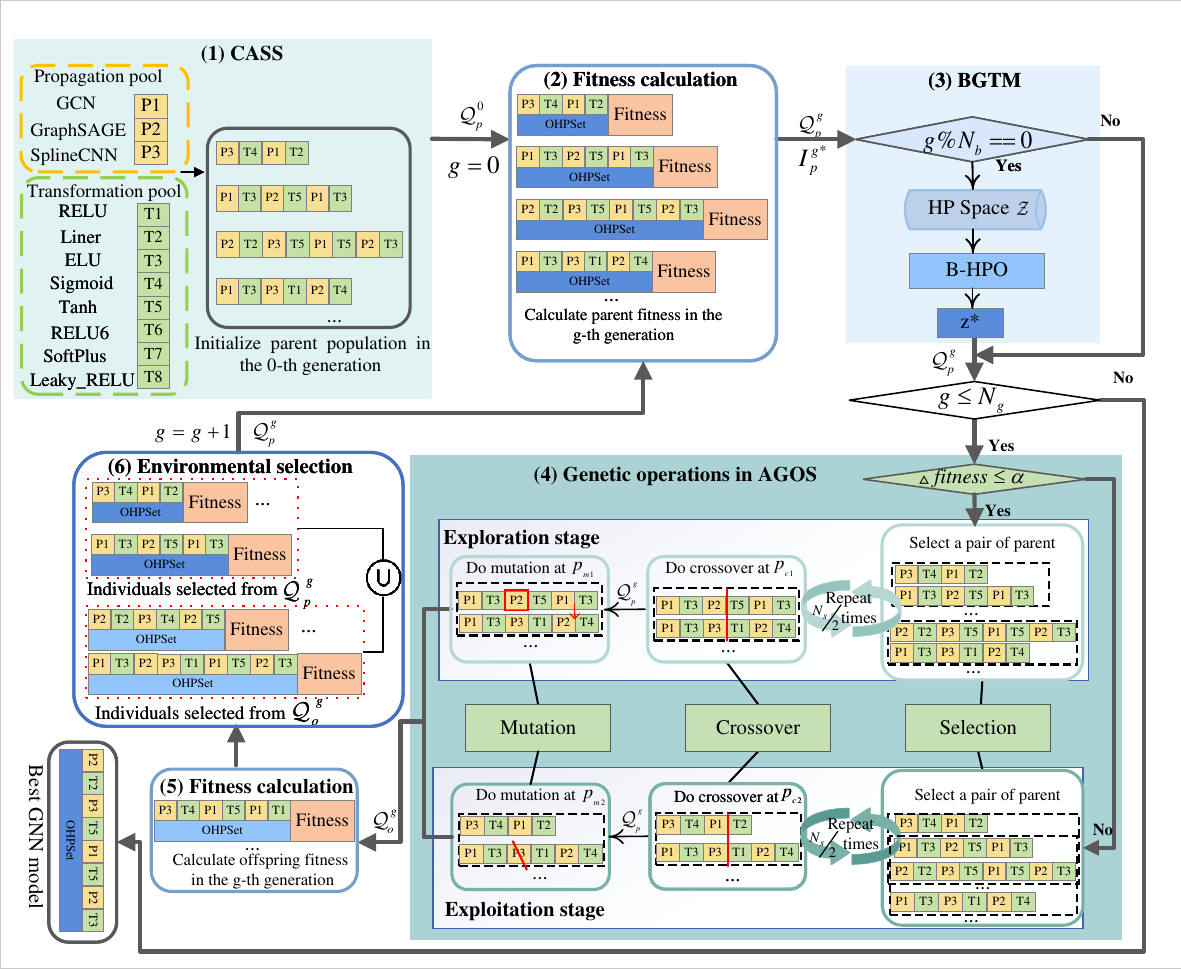}  
     \caption{Overview of the ABG-NAS framework: (1) An initial population $\mathcal{Q}_{p}^g$ = $\mathcal{Q}_{p}^0$ is generated from the expanded search space; (2) Fitness evaluation is conducted for the GNN architectures in the parent population $\mathcal{Q}_{p}^g$, which includes the individual \( I_p^{g*} \) with the highest fitness value; (3) The BGTM algorithm performs hyperparameter optimization in every \(N_b\) generations thereafter; (4) The genetic operations in AGOS algorithm guide individuals go through exploration and exploitation phases based on fitness, performing selection, crossover, and mutation to generate offspring ${Q}_{o}^g$ from ${Q}_{p}^g$; (5) Offspring fitness is calculated on ${Q}_{o}^g$; (6) The next generation is formed by selecting individuals from both the parent population \({Q}_{p}^g\) and the offspring population \({Q}_{o}^g\) based on their fitness values. Then return to (3) to update the fitness values of the next generation ${Q}_{p}^g$; If $g\%N_b$==0, return to (3). Otherwise, if $g \leq G$, return to (4) and if $g==G$, output the best individual in ${Q}_{p}^G$.}  
    \label{fig:framework}
\end{figure}

In this section, we present a detailed introduction to the ABG-NAS framework. Figure \ref{fig:framework} depicts the overall structure of ABG-NAS, highlighting its key components and workflow. The subsequent sections will delve into the design specifics and provide corresponding pseudocode, offering a comprehensive explanation of how ABG-NAS effectively searches for and optimizes GNN architectures.

\subsection{Problem Formulation}
\label{subsec:problem_formulation}

DFG-NAS \cite{zhang2022dfg} formulated the GCN NAS problem as a bi-level optimization task, involving outer-level GNN architecture optimization subjected to the inner-level GNN model parameter optimization, as formulated in Equations (\ref{eq:outer_level}) and (\ref{eq:inner_level}).  

\begin{align}
I^{*} &= \arg\max_{I\in \mathcal{A}} \mathbb{E}_{v_i \in V_{val}} \phi\left( y_i, f_{I}(\mathbf{x}_i, \mathbf{e}_i;z,\Theta^*)\right), 
\label{eq:outer_level} \\
\text{s.t.} \quad 
\Theta^* &= \arg\min_{\Theta}  \mathbb{E}_{v_i \in V_{train}} \ell \left[  y_i,f_{I}(\mathbf{x}_i, \mathbf{e}_i;z,\Theta)\right]. 
\label{eq:inner_level}
\end{align}

Here, \(v_i\) represents a node from the training set \(V_{train}\) or the validation set \(V_{val}\). \(I^{*}\) and \(\Theta^*\) denote the optimized GNN architecture selected from the search space \(\mathcal{A}\) and the set of optimized model parameters for a GNN architecture, respectively. \(y_i\) is the true label of node \(v_i\), and \(\mathbf{x}_i\) is its feature vector. \(\mathbf{e}_i = \{e_{ij}\}_{j=1}^{N}\) is the set of edges associated with \(v_i\). \(f_{I}\) is the mapping function of the GNN model decoded from an architecture \(I\) and $z$ is the combination of hyperparameters of the GNN model. \(\hat{y}_i = f_{I}(\mathbf{x}_i, \mathbf{e}_i;z,\Theta)\) is the predicted label of node \(v_i\). \(\ell(\cdot)\) is the loss function used to optimize the model parameters \(\Theta\). \(\phi(\cdot)\) is the fitness function used in a GA process to evaluate the performance of an architecture \(I\).

In DFG-NAS, $z$ is set based on expert knowledge. To improve the overall performance of $I^*$ by optimize the hyperparameters of GNN models, we further formulate a hyperparameter optimization process as Equation (\ref{eq:z_level}) after obtaining $\Theta^*$ and $I^*$. 

\begin{equation}
z^* = \arg\max_{z\in \mathcal{Z}} \mathbb{E}_{v_i \in V_{val}}\psi\left[ y_i, f_{I^*}(\mathbf{x}_i, \mathbf{e}_i;z,\Theta^*)\right] 
\label{eq:z_level} \\
\end{equation}

Here, $\psi(\cdot)$ is an objective function employed to evaluate the performance of $z\in \mathcal{Z}$. The hyperparameter combination search space \( \mathcal{Z} \) is defined in Equation \ref{eq:search_space}.

\begin{align}
\mathcal{Z} = \left\{ z = \langle \beta_i \rangle_{i=1}^{N_z} \right\},
\label{eq:Z_search_space}
\end{align}
where \( \beta_i \in \mathbb{B}_i \) is a single hyperparameter of a GNN model, such as hidden layer dimensions, dropout rates, learning rate, and weight decay coefficient. $\mathbb{B}_i$ is the search space of $\beta_i$. ${N_z}$ is the total number of hyperparameters in the to-be optimized the hyperparameter combination. These hyperparameters are crucial for controlling the training process and improving model performance. For example, the learning rate determines the speed of model convergence, while the dropout rate helps prevent overfitting.

\subsection{Comprehensive Architecture Search Space}
\label{subsec:flexible-search-space}

ABG-NAS introduces a more comprehensive architecture search space by expanding the diversity of \textbf{P} and \textbf{T} operations, enabling flexible combinations of components tailored for GNN neural architecture search (NAS). This design addresses the limitations of prior methods, which often restrict the search space to fixed \textbf{P} and \textbf{T} operations. It further enhances the search space by incorporating various internal configurations of the \textbf{P} and \textbf{T} operations. As illustrated in Figure \ref{fig:framework}, eight configurations for the \textbf{T} operation are included, such as ReLU, Linear, and Sigmoid, thereby enabling a broader exploration of potential architectures. The expanded search space (ESS) \( \mathcal{A} \) is a set containing all individual \( I \) that is mathematically defined by Equation (\ref{eq:search_space}).

\begin{align}
\mathcal{A} = \left\{ I = \langle c_i \rangle_{i=1}^d \mid d \in [d_{\text{min}}, d_{\text{max}}] \cap \mathbb{Z}^+, \ c_i \in \mathcal{O} \right\},
\label{eq:search_space}
\end{align}
where \( I \) denotes an individual architecture within the population, \( d \) represents a flexible length of the sequence of operations, and \( c_i \) refers to the \( i \)-th operation in the sequence, selected from the operation set \( \mathcal{O} \) defined as Equation (\ref{eq:operation_Set}).

\begin{equation}
\mathcal{O}=\left\{\{\textbf{P}_i\}_{i=1}^{N_P},\{\textbf{T}_j\}_{j=1}^{N_T}\right \}  
\label{eq:operation_Set}
\end{equation}
where \( N_P \) and \( N_T \) are the counts of available propagation and transformation operations, respectively. This flexible search space allows for diverse combinations of \textbf{P} and \textbf{T}, facilitating adaptation to various graph tasks and enabling the exploration of a wide array of potential GNN configurations.

\subsection{Adaptive Genetic Optimization Strategy }
\label{subsec:AGOS}

AGOS enhances genetic optimization in GNAS by introducing adaptive evolutionary dynamics and diverse mating operations to accelerate convergence and improve search efficiency. It operates through seven key steps, featuring an adaptive genetic operation switching mechanism, detailed as follows:

\textbf{Step 1: Population Initialization.} The initial parent population of size \( N_s \) is randomly generated from the search space defined by Equation (\ref{eq:search_space}), represented as \( \mathcal{Q}_{p}^0 = \{ I_{pi}^0 \}_{i=1}^{N_s} \). Compared to DFG-NAS, the operation set \( \mathcal{O} \) has been expanded to include a broader range of options, as specified in Equation (\ref{eq:operation_Set}).

\textbf{Step 2: Fitness Calculation.} Similar to DFG-NAS, the fitness scores for generation \( g \), denoted as \( \mathcal{S}_{p}^{g} = \{ s_{pi}^{g} \}_{i=1}^{N_s} \), are calculated by decoding each individual \( I_{pi}^g \in \mathcal{Q}_{p}^g \) into a corresponding GNN architecture, training the model on the training set, and evaluating its performance on the validation set. However, unlike DFG-NAS, which uses accuracy as the fitness function, EAGA employs the macro F1 score to better address class imbalance issues that are prevalent in real-world node classification tasks. Here, \( s_{pi}^{g} \) represents the fitness score of the \( i \)-th parent individual.

\textbf{Step 3: Adaptive Genetic Operation Switching.}
In genetic algorithms, selection and mating (including crossover and mutation) operations are key genetic operations that drive the evolutionary process. However, in DFG-NAS, as outlined in Section \ref{subsec:EA-background} Steps 3-5, these operations are applied uniformly with fixed strategies, regardless of fitness score improvements throughout the search process. This fixed approach has several drawbacks, including a higher risk of premature convergence and inefficiency in navigating the solution space due to the lack of differentiation between exploration and exploitation phases.

Inspired by the concepts of exploration and exploitation in evolutionary algorithms \cite{liu2021survey}, we incorporate these principles into the GNAS problem by designing adaptive genetic operations. In evolutionary algorithms, exploration aims to search the solution space broadly, generating diverse solutions to avoid local optima, while exploitation focuses on refining existing solutions to converge on the optimal one. The proposed AGOS adaptively alternates between these two stages based on fitness improvement, denoted as \(\Delta \text{fitness}(g)\). When \(g = 0\), \(\Delta \text{fitness}(g) = 0\); otherwise, \(\Delta \text{fitness}(g)\) is calculated as shown in Equation (\ref{eq:delta_fitness}).

\begin{equation}
\label{eq:delta_fitness}
\Delta \text{fitness}(g) = \lambda \times \overline{S_p^g} + (1 - \lambda) \times \Delta \text{fitness}(g - 1)
\end{equation}

Here, \(\lambda\) is a smoothing factor between 0 and 1 that determines the influence of \(\overline{s_p^g}\) on \(\Delta \text{fitness}(g)\), where \(\overline{s_p^g}\) represents the current generation's average fitness, calculated by Equation (\ref{eq:avg_fitness}).

\begin{equation}
\label{eq:avg_fitness}
\overline{s_p^g} = \frac{1}{N_s} \sum_{i=1}^{N_s} s_{p_i}^g,
\end{equation}
where \(N_s\) denotes the population size, and \(s_{p_i}^g\) is the fitness score of the \(i\)-th individual in generation \(g\).

As illustrated in Figure \ref{fig:framework} (4) Genetic operations in AGOS, when \(\Delta \text{fitness}(g)\) is no larger than a predefined threshold \(\alpha\), the algorithm enters the exploration stage; otherwise, it enters the exploitation stage. The main distinction between exploration and exploitation in AGOS lies in the values of genetic operation control factors, i.e., the selection intensity \(k\), crossover probability \(p_{c}\), and mutation probability \(p_{m}\). During the exploration phase, higher values (denoted as \(k_1, p_{c_1}, p_{m_1}\)) are used to increase diversity and broaden the search space. Conversely, during the exploitation phase, lower values (denoted as \(k_2, p_{c_2}, p_{m_2}\)) are employed to focus on refining high-quality solutions and accelerating convergence, as demonstrated by Equation (\ref{eq:adaptive_switch}).

\begin{align}
\label{eq:adaptive_switch}
k, p_{c}, p_{m} =
\begin{cases}
k_1, p_{c_1}, p_{m_1}, & \Delta \text{fitness}(g) \leq \alpha, \\
k_2, p_{c_2}, p_{m_2}, & \text{otherwise}.
\end{cases}
\end{align}

This adaptive genetic operation switching mechanism allows AGOS to effectively navigate between exploration and exploitation stages, enhancing adaptability and efficiency in finding optimal GNN architectures.

\textbf{Step 4: Selection}:  In this step, a parent pair is selected from the population $\mathcal{Q}_p^g$ for each crossover. This selection process is repeated $N_s / 2$ times to maintain the population size. A \(k\)-tournament selection method is used to select each parent, where \(k\) individuals are randomly selected from $\mathcal{Q}_p^g$, and the one with the highest fitness score is chosen as a parent. The value of \(k\) is adaptively determined by Equation (\ref{eq:adaptive_switch}).% The process is demonstrated in Figure \ref{fig:cro&mu}.

\begin{figure}[ht]
\centering
\includegraphics[width=\linewidth,clip,trim=5 5 5 5]{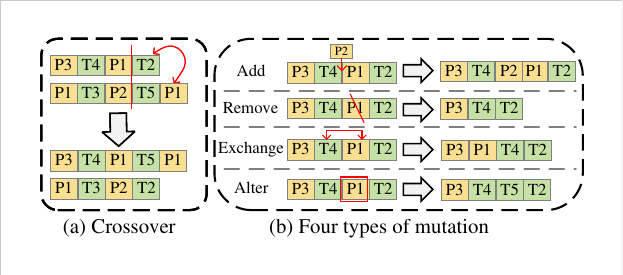} 
\caption{(a) Single-point crossover where segments between parents are exchanged. (b) Four types of mutation operations: Add, Remove, Exchange, and Alter.}
\label{fig:cro&mu}
\end{figure}

\textbf{Step 5: Mating Operations.} As the first GA framework based on \textbf{P} and \textbf{T} operations, DFG-NAS simply performs mutation by selecting a single parent for either adding or removing operations. We extend their work by selecting two parents and introducing a single-point crossover, along with additional exchange and alter mutations.

\textbf{Crossover.} In our framework, selected parent pairs undergo a single-point crossover with a probability \( p_{c} \) determined by Equation (\ref{eq:adaptive_switch}), which also repeats $N_s/2$ times followed by the selection operation.  During a single-point crossover, each parent is split into two segments at a randomly selected position, and the second segments of both parents are exchanged with each other. For example, as illustrated in Figure \ref{fig:cro&mu} (a), the first parent \textbf{P3-T4-P1-T2} and the second parent \textbf{P1-T3-P2-T5-P1} are split at the position marked by the red vertical line. The second segments are swapped, forming new individuals: \textbf{P3-T4-P1-T5-P1} and \textbf{P1-T3-P2-T2}. Here, \textbf{\{P$_i$\}}$_{i=1}^3$ and \textbf{\{T$_i$\}}$_{i=1}^5$ represent different variants of the \textbf{P} and \textbf{T} operations from the set \(\mathcal{O}\).

\textbf{Mutation.} After selection and crossover, each individual in the current generation undergoes mutation with a probability \( p_m \), which determines whether a mutation is applied. If a mutation occurs, one of four mutation types—add, remove, exchange, or alter—is randomly selected, as shown in Figure \ref{fig:cro&mu} (b). \textbf{Add} means a random operation from \(\mathcal{O}\) is inserted into a randomly selected position in the individual. For example, the individual \textbf{P3-T4-P1-T2} has \textbf{P2} inserted at the second position, resulting in the new individual \textbf{P3-T4-P2-P1-T2}. \textbf{Remove} represents that a randomly selected operation is deleted from the individual. For instance, in the individual \textbf{P3-T4-P1-T2}, the operation \textbf{P1} is removed, forming the new individual \textbf{P3-T4-T2}. \textbf{Exchange} is the operation in which two randomly chosen operations within the individual are swapped. For example, in the individual \textbf{P3-T4-P1-T2}, the operations \textbf{T4} and \textbf{P1} are exchanged, resulting in the new individual \textbf{P3-P1-T4-T2}.  \textbf{Alter} presents a randomly selected operation that is replaced by another operation from \(\mathcal{O}\).  For instance, in the individual \textbf{P3-T4-P1-T2}, the operation \textbf{P1} is altered to \textbf{T5}, forming the new individual \textbf{P3-T4-T5-T2}.

These mutations introduce variation into the population, preventing premature convergence.

\textbf{Step 6: Fitness Calculation.} After mutation, the offspring population for generation \( g \), denoted as \( \mathcal{Q}_o^g \), is generated. To assess the performance of each individual in \( \mathcal{Q}_o^g \), fitness scores are calculated. The fitness scores for the offspring in generation \( g \) are represented as \( \mathcal{S}_{o}^{g} = \{ s_{oi}^{g} \}_{i=1}^{N_s} \), where \( s_{oi}^{g} \) denotes the fitness score of the \( i \)-th offspring.

\textbf{Step 7: Environmental Selection.} This step maintains the fixed population size \( N_s \) by selecting individuals from the union of the parent and offspring populations, denoted as \( \mathcal{Q}^{g} = \mathcal{Q}_{p}^{g} \cup \mathcal{Q}_{o}^{g} \). By selecting from both populations, AGOS preserves genetic diversity, reducing the risk of premature convergence and getting stuck in local optima. Specifically, an elitism roulette selection mechanism is employed to generate the next parent population. The individual with the highest fitness score, \( I^{g*} \in \mathcal{Q}^{g} \), is guaranteed to be included in \( \mathcal{Q}_{p}^{g+1} \), while the remaining \( N_s - 1 \) individuals in \( \mathcal{Q}_{p}^{g+1} \) are selected using the roulette selection mechanism. This approach preserves the strongest individual through elitism while ensuring diversity through roulette selection, allowing AGOS to efficiently explore and exploit the search space, thereby increasing the chances of finding the optimal solution.

Environmental selection concludes the current generation \( g \) and prepares the population for the next generation \( g = g+1 \), setting up the algorithm for the next iteration. If \( g \leq N_g \), the algorithm returns to Step 2 to proceed to the next generation; otherwise, if \( g = N_g  \), the evolution concludes, and \( I^{g*} \) is returned as the optimal GNN architecture.

\subsection{Bayesian-Guided Tuning Module}
\label{subsec:phased-bayesian-optimization}

In the GNN architecture search process, adjusting training hyperparameters, such as hidden layer size and learning rate, significantly impacts model performance and convergence speed. However, our baseline method, DFG-NAS, mainly focuses on architecture search while neglecting the optimization of training hyperparameters, potentially leaving the model's potential underutilized and affecting both search efficiency and final performance.

To address this issue, we propose a BGTM strategy to periodically embed a Bayesian HPO (B-HPO) algorithm into the AGOS G-NAS process based on a preset interval $N_b$.  This strategy achieves a balance between hyperparameter tuning and architecture search, particularly exhibiting significant dual optimization effects on complex, large-scale datasets.

Both the BGTM algorithm for hyperparameter optimization and the AGOS process for G-NAS are crucial components of the proposed ABG-NAS framework, as illustrated in Figure \ref{fig:framework}. In the rest of this section, we further present the implementation detail of BGTM and AGOS within the pseudocode of the whole ABG-NAS framework, as shown in Algorithm \ref{alg:abgnas}.

\begin{figure}[H]
\centering
\begin{minipage}{\linewidth}
\begin{algorithm}[H]
\caption{Procedure of ABG-NAS}
\label{alg:abgnas}
\textbf{Input:} $\mathcal{A}$, $\mathcal{Z}$, $N_s$, $N_g$, $N_b$, \(\phi(\cdot)\), $\alpha$, \(k_1, p_{c_1}, p_{m_1}\), \(k_2, p_{c_2}, p_{m_2}\). 

\textbf{Output:} $I^*$, $z^*$ %, $z^*$, $\Theta^*$.

\begin{algorithmic}[1] 
\STATE g = 0, $\Delta \text{fitness}(0)$=0, and initialize $\mathcal{Q}_{p}^0  = \{ I_{pi}^0 \}_{i=1}^{N_s} $ from $\mathcal{A}$.

\STATE Identify the optimal parent individual \( I_p^{0*} \in \mathcal{Q}_{p}^0 \) by solving Equations (\ref{eq:outer_level}) and (\ref{eq:inner_level}), and calculate fitness scores, \( \mathcal{S}_{p}^{0}\), for $\mathcal{Q}_{p}^0$ based on \(\phi(\cdot)\). 

\STATE Determine the optimal hyperparameter combination \( z^{*} \in \mathcal{Z} \) for \( I_p^{0*} \) by solving Equation (\ref{eq:z_level}) using B-HPO.

\FOR{ $g \in [0,N_g]\cap \mathbb{Z}^+$}

\STATE Calculate $\Delta \text{fitness}(g)$ based  on Equation \ref{eq:delta_fitness}) if $g>0$.

\STATE Set \(k\), \(p_{c}\), and \(p_{m}\) based on Equation (\ref{eq:adaptive_switch}).

\IF{$\Delta \text{fitness}(g) \leq \alpha$ }

\STATE Enter the exploration stage of AGOS and execute selection, crossover, and mutation with \(k_1, p_{c_1}, p_{m_1}\) to generate $\mathcal{Q}_{o}^g$.

\ELSE 

\STATE Enter the exploitation stage of AGOS and execute selection, crossover, and mutation with \(k_2, p_{c_2}, p_{m_2}\) to generate $\mathcal{Q}_{o}^g$.

\ENDIF 

\STATE Calculate fitness scores, \({S}_{o}^{g}\), for $\mathcal{Q}_{o}^g$ based on \(\phi(\cdot)\). %= \{ s_{oi}^{g} \}_{i=1}^{N_s}

\STATE Execute environmental selection to generate next parent generation $\mathcal{Q}_{p}^{g}$ from \( \mathcal{Q}^{g} = \mathcal{Q}_{p}^{g} \cup \mathcal{Q}_{o}^{g} \).

\STATE $g=g+1$

\STATE Identify the optimal parent individual \( I_p^{g*} \in \mathcal{Q}_{p}^g \) by solving Equations (\ref{eq:outer_level}) and (\ref{eq:inner_level}), and calculate fitness scores, \( \mathcal{S}_{p}^{g}\), for $\mathcal{Q}_{p}^g$ based on \(\phi(\cdot)\). 

\IF{$g \% N_b == 0$}

\STATE Find out $I_p^{g*} \in \mathcal{Q}_{p}^g$ by solving Equations (\ref{eq:outer_level}) and (\ref{eq:inner_level}).

\STATE Find out $z^{*} $ for $I_p^{g*}$ by solving Equation (\ref{eq:z_level})  using B-HPO.
\ENDIF

\ENDFOR

\RETURN $I^*$ = $I_p^{g*}$, $z^*$. 

\end{algorithmic}
\end{algorithm}
\end{minipage}
\end{figure}

The inputs of this ABG-NAS include the GNN search space, $\mathcal{A}$, defined by Equation (\ref{eq:search_space}); the hyperparameter combination search space, $\mathcal{Z}$, defined by Equation (\ref{eq:Z_search_space}); the population size of AGOS algorithm $N_s$; the maximum iteration generation $N_g$; the fitness function defined for AGOS algorithm \(\phi(\cdot)\); the PB-HBO iteration interval $N_b$; a predefined threshold, $\alpha$, for distinguishing exploration and exploitation stages of AGOS; and the predefined selection intensity, crossover probability, and mutation probability for both exploration and exploitation stages, \(\{k_1, p_{c_1}, p_{m_1}\}\), and \(\{k_2, p_{c_2}, p_{m_2}\}\). To ensure the last integration executes a B-HPO process, set $N_g$ to be an integer multiple of $N_b$. The outputs of ABG-NAS are the optimized GNN architecture $I^*$ with optimized hyperparameter combination $z^*$. 

As shown in Step 1, ABG-NAS starts with setting the current generation $g=0$, $\Delta \text{fitness}(0)$=0, and initialize $\mathcal{Q}_{p}^0  = \{ I_{pi}^0 \}_{i=1}^{N_s} $ from $\mathcal{A}$.  

Then, in Step 2, the optimal parent individual \( I_p^{0*} \in \mathcal{Q}_{p}^0 \) is identified by solving Equations (\ref{eq:outer_level}) and (\ref{eq:inner_level}), and the fitness scores, \( \mathcal{S}_{p}^{0}\), is calculated for $\mathcal{Q}_{p}^0$ with the fitness function \(\phi(\cdot)\), as demonstrated in Figure\ref{fig:framework} (2) Fitness calculation.   

In Step 3, a B-HPO algorithm executes to determine the optimal hyperparameter combination \( z^{*} \in \mathcal{Z} \) for \( I_p^{0*} \) by solving Equation (\ref{eq:z_level}), as demonstrated in Figure\ref{fig:framework} (3) BGTM algorithm. 

Then, in Steps 4 to 15, ABG-NAS executes the AGOS algorithm iteratively to iditify the best GNN architecture $I_p^{g*}$. Specifically, Steps 5 and 6 control the adative genetic operation switching based on the value of $\Delta \text{fitness}(g)$. Then, in Steps 7 to 11, AGOS enters either the exploration or the exploitation stage to do the selection, crossover and mutation genetic operations to generate the offspring of the current generation, $\mathcal{Q}_{o}^g$. Next, Steps 12 and 14 evaluate the offspring and generate the next generation's parent population form both $\mathcal{Q}_{o}^g$ and $\mathcal{Q}_{p}^g$. Lastly, the best parent individual \( I_p^{g*} \in \mathcal{Q}_{p}^g \) for the current generation is identified by solving Equations (\ref{eq:outer_level}) and (\ref{eq:inner_level}), and the corresponding scores, \( \mathcal{S}_{p}^{g}\), for $\mathcal{Q}_{p}^g$ are calculated.  

Subsequently, in Steps 16 to 19, the B-HPO algorithm is periodically executed with an interval $N_b$ to determine the optimized hyperparameter combination $z^*$ for \( I_p^{g*} \).

Finally, when the current generation $g$ reaches the maximum generation $N_g$, $I^* = I_p^{g*}$ and $z^*$ are returned as the output of ABG-NAS.

In summary, the ABG-NAS framework provides a comprehensive multi-level optimization strategy, balancing architecture exploration and hyperparameter tuning.  We validate the performance of ABG-NAS through extensive experiments in the following section.

\section{Experimental Analysis}
\label{sec:exp}
This section presents the experimental analysis of ABG-NAS, beginning with the experimental settings and dataset descriptions. Next, we evaluate ABG-NAS by comparing it against existing experience-based GNN architectures, followed by a comparison with state-of-the-art EA-based NAS methods, including DFG-NAS. Finally, we conduct ablation studies to investigate the contributions of key components within the ABG-NAS framework, demonstrating the effectiveness of each component in the overall performance.

\subsection{Experimental Settings}

\subsubsection{Datasets Details}
We evaluate the proposed ABG-NAS algorithm on four widely-used public datasets: Cora\cite{mccallum2000efficient}, PubMed \cite{sen2008collective}, CoraFull \cite{bojchevski2017deep}, and Citeseer\cite{sen2008collective}.  These datasets differ significantly in size, sparsity, and structural complexity, enabling a comprehensive evaluation of ABG-NAS's robustness and adaptability across diverse graph structures. Cora and PubMed are citation networks, while CoraFull is an extended version of Cora with significantly higher connectivity. In contrast, Citeseer is characterized by its higher sparsity, including 217 isolated nodes, making it particularly challenging for GNN models to extract meaningful patterns in fragmented graph data. Each dataset is split into training, validation, and testing sets in a 6:2:2 ratio. Refer to Table~\ref{tab:table1_dataset_information} for detailed dataset statistics.

\begin{table*}[ht]
\centering
\caption{Summary of dataset statistics used in the experiments. D\textsubscript{in} and D\textsubscript{out} represent the average in-degree and out-degree, respectively, while D\textsubscript{avg} denotes the average degree.}
\label{tab:table1_dataset_information}
\resizebox{\textwidth}{!}{
% \begin{tabular}{lcccccccc}
\begin{tabular}{cccccccccc}
\toprule
Dataset & Nodes & Isolated Nodes & Edges & Classes & D\textsubscript{in} & D\textsubscript{out} & D\textsubscript{avg} & Connected Nodes \\ \midrule
Cora\cite{mccallum2000efficient}     & 2,708  & 0   & 5,278   & 7    & 1.95  & 1.95  & 3.90   & 2,708 \\ 
PubMed\cite{sen2008collective}       & 19,717 & 0   & 44,338  & 3    & 2.25  & 2.25  & 4.50   & 19,717 \\
CoraFull\cite{bojchevski2017deep}    & 19,793 & 0   & 126,842 & 70   & 6.41  & 6.41  & 12.82  & 19,793 \\ 
Citeseer\cite{sen2008collective}     & 3,327  & 217 & 4,732   & 6    & 1.42  & 1.42  & 2.84   & 3,110 \\
\bottomrule
\end{tabular}
}

\end{table*}

\subsubsection{Parameter Settings}
\label{subsec:parameterSetting}

The ABG-NAS framework was configured with specific parameters to efficiently guide the GNN architecture search. For the expanded search space, the operation set \( \mathcal{O} \) consisted of 5 \textbf{P} operations: GCN, SplineCNN, and three GraphSAGE variants (\textit{mean}, \textit{max}, and \textit{sum}), as well as 8 \textbf{T} operations: \textit{ReLU}, \textit{Linear}, \textit{ELU}, \textit{Sigmoid}, \textit{Tanh}, \textit{ReLU6}, \textit{SoftPlus}, and \textit{Leaky\_ReLU}. GCN was selected for its simplicity and efficiency in feature aggregation, while GraphSAGE and SplineCNN offer alternative aggregation and transformation mechanisms.

For the AGOS algorithm, the depth \( d \) of each individual in the population was set to range from 3 to 15. Both the population size \( N_s \) and the maximum number of generations \( N_g \) were set to 20. An early stopping strategy with a maximum of 300 epochs was used to train the GNN models for each individual.

To ensure fair and robust comparisons, all experiments were repeated across 10 independent runs using different random seeds. Additionally, no learning rate scheduler was applied for any method. A fixed learning rate was used throughout training to avoid scheduler-induced bias and maintain consistency across all baselines and the proposed framework.

In the Bayesian-Guided Tuning Module (BGTM), the update interval \( N_b \) was empirically set to 5 generations, balancing tuning benefits and computational overhead.

The hyperparameters optimized by BGTM and their search spaces were defined as follows: the hidden dimension size was set between \( 4 \) and \( 256 \), forward dropout rate was varied between \( 0.4 \) and \( 0.6 \), middle dropout rate was tuned between \( 0.2 \) and \( 0.4 \), the overall dropout rate was adjusted in the range \( 0.3 \) to \( 0.5 \), the learning rate was adjusted in the range \( 10^{-4} \) to \( 10^{-1} \), and weight decay was tuned from \( 10^{-5} \) to \( 10^{-2} \).

In addition, the smoothing factor \( \lambda = 0.5 \) used in the AGOS fitness control mechanism (Equation~\ref{eq:delta_fitness}) was fixed to provide balanced influence between current and historical fitness trends. This value was empirically selected based on preliminary experiments for stability.

All methods were executed under the same computational budget, with 10 independent runs per dataset and consistent train/validation/test splits to ensure a fair comparison. The reported results are averaged over 10 runs, and standard deviations are included to indicate performance stability.

\subsubsection{Compared Algorithms}

\textbf{Compared Algorithms.} To ensure a fair and reproducible comparison, we provide below the configuration details of all baseline NAS methods evaluated in our experiments:

\textbf{GraphNAS} and \textbf{SNAG}: These reinforcement learning-based NAS methods were executed using their official implementations with default configurations. As these models are computationally intensive, we reused the results reported in their original papers (as marked with *).

\textbf{SANE} and \textbf{Auto-HeG}: These gradient-based methods were re-implemented using the authors’ official codebases. We retained all default settings without additional hyperparameter tuning.

\textbf{DE}: The differential evolution baseline was configured using standard defaults: scaling factor \( F = 0.8 \) and crossover rate \( CR = 0.9 \).

\textbf{PSO}: The inertia weight was fixed at \( w = 0.5 \), while cognitive and social coefficients were set to \( c_1 = 1.5 \) and \( c_2 = 1.5 \), respectively. The velocity was bounded in \([-1, 1]\). Coefficients \( c_1 \) and \( c_2 \) were selected via grid search within \([1.0, 2.0]\).

\textbf{DFG-NAS}: The mutation probability was set to 0.3, and tournament selection size to 3, in accordance with the original implementation.

\subsection{Comparison with Experience-based GNN Architectures}
\label{subsec:comparison_manual_gnn}

We first compare the performance of the AGOS-driven GNAS process with experience-based GNN architectures on three GNN models: GCN, SplineCNN, and GraphSAGE. The results are summarized in Table \ref{tab:table2_manual_Comparison}. In the ``Methods" column, EXP refers to a classical two-layer \textbf{P-T-P-T} architecture manually designed based on expert knowledge, where the \textbf{P} and \textbf{T} operations are derived from the base model shown in the first column. AGOS denotes the use of the proposed AGOS algorithm to automate the architecture search within the same operation set as the corresponding EXP baseline. The final row shows ABG-NAS, which operates on the expanded search space as defined in Subsection \ref{subsec:parameterSetting}.

\begin{table*}[h]
\centering
\caption{Comparison of accuracy (Acc) and macro F1 scores between experience-based (EXP) GNN architectures, AGOS-automated architectures, and ABG-NAS across four datasets. The base models include GCN, SplineCNN, and GraphSAGE. The table reports the mean accuracy and macro-F1 score over 10 independent runs, with standard deviations included. }
\label{tab:table2_manual_Comparison}
\scriptsize % 
\setlength{\tabcolsep}{4pt} 
\renewcommand{\arraystretch}{0.9} 
\resizebox{\textwidth}{!}{
\begin{tabular}{ccccccccccccc}
\toprule
\multirow{2}{*}{Base Model} & \multirow{2}{*}{Methods }& \multicolumn{2}{c}{Cora\cite{mccallum2000efficient}} &  & \multicolumn{2}{c}{PubMed \cite{sen2008collective}} &  & \multicolumn{2}{c}{CoraFull \cite{bojchevski2017deep}} &  & \multicolumn{2}{c}{Citeseer\cite{sen2008collective}} \\ \cmidrule(lr){3-4} \cmidrule(lr){6-7} \cmidrule(lr){9-10} \cmidrule(lr){12-13} 
\multicolumn{2}{c}{} & \scriptsize Acc (\%) & \scriptsize Macro-F1 (\%) &  & \scriptsize Acc (\%) & \scriptsize Macro-F1 (\%) &  & \scriptsize Acc (\%) & \scriptsize Macro-F1 (\%) &  & \scriptsize Acc (\%) & \scriptsize Macro-F1 (\%) \\ \midrule
\multirow{2}{*}{GCN \cite{kipf2016semi}} & EXP & 86.76±0.07 & 85.65±0.06 &  & 86.57±0.01 & 86.98±0.03 &  & 62.66±0.06 & 54.28±0.09 &  & 72.41 ±0.03 & 72.78 ±0.04 \\
 & AGOS & 87.93±0.02 & 86.40±0.01 &  & 89.05±0.03 & 88.82±0.09 &  & 64.25±0.03 & 55.06±0.01 &  & 72.89±0.02 & 73.93 ±0.06 \\
\multirow{2}{*}{SplineCNN \cite{fey2018splinecnn}} & EXP & 86.86±0.05 & 85.78±0.03 &  & 88.22±0.06 & 88.08±0.03 &  & 57.34±0.06 & 48.62±0.05 &  & 71.06±0.07 & 70.69±0.08 \\
 & AGOS & 86.51±0.04 & 85.99±0.02 &  & 89.23±0.03 & 89.37±0.04 &  & 65.02±0.07 & 54.36±0.09 &  & 73.11±0.02 & 73.52±0.04 \\
\multirow{2}{*}{GraphSAGE \cite{hamilton2017inductive}} & EXP & 86.76±0.03 & 85.58±0.04 &  & 88.22±0.07 & 88.01±0.01 &  & 56.95±0.01 & 48.22 ±0.06 &  & 71.20 ±0.05 & 70.85±0.05 \\
 & AGOS & 86.88±0.05 & 85.67±0.02 &  & 89.64±0.03 & 89.64±0.02 &  & 65.34±0.05 & 55.18±0.09 &  & 72.76±0.07 & 73.41±0.03 \\\midrule
 $\mathcal{O}$ & ABG-NAS & \textbf{88.32±0.02} &  \textbf{87.52±0.05} &  & \textbf{89.88±0.04} & \textbf{89.67±0.02} &  & \textbf{72.14±0.06} & \textbf{63.03±0.07} &  & \textbf{77.44±0.02} & \textbf{73.50±0.01} \\ \bottomrule
\end{tabular}}

\end{table*}

The results in Table \ref{tab:table2_manual_Comparison} demonstrate the clear advantages of the automated AGOS approach over manual GNN designs across all three base models and four datasets. For instance, on the Cora dataset, GCN’s accuracy improved from 86.76\% to 87.93\%, with the F1 score increasing from 85.65\% to 86.40\%. On PubMed, the accuracy rose from 86.57\% to 89.05\%, while the F1 score improved from 86.98\% to 88.82\%. These improvements indicate that the automated NAS process consistently discovers more effective architectures, adapting better to different datasets. ABG-NAS, leveraging a broader and more flexible search space, outperforms all models, achieving the best performance across all datasets, particularly excelling on larger datasets like CoraFull and Citeseer.

\textbf{Impact of Dataset Characteristics.}
While the automated AGOS NAS method consistently outperforms the experience-based (EXP) methods across all datasets, the degree of improvement is closely related to the characteristics of the datasets themselves. As shown in Table \ref{tab:table1_dataset_information}, the Cora dataset is the smallest, with only 2,708 nodes and 5,278 edges. The gains achieved by AGOS on Cora are relatively modest compared to larger and more complex datasets like PubMed and CoraFull. For instance, the accuracy improvement for GCN on Cora is 1.17\% (87.93\%-86.76\%), whereas on PubMed and CoraFull, the improvements are 2.48\% (89.05\%-86.57\%) and 1.59\% (64.25\%-62.66\%), respectively. These substantial improvements on PubMed (19,717 nodes, 44,338 edges) and CoraFull (19,793 nodes, 126,842 edges) demonstrate that AGOS is particularly effective on larger, more complex graphs. This is because the flexibility in architectural design provided by NAS methods allows for the creation of sophisticated models that can effectively capture both local and global patterns.

Notably, ABG-NAS also achieves impressive performance on the sparse Citeseer dataset, with a 4.03\% accuracy improvement on the GCN base model (77.44\%-72.41\%), which is higher than the improvement on the denser Cora dataset (1.17\%, 87.93\%-86.76\%). Citeseer’s low average degree (\(D_{\text{avg}} = 2.84\)) and 217 isolated nodes pose significant challenges for GNNs to propagate information effectively. However, the adaptive architectures generated by ABG-NAS are able to mitigate some of the inherent limitations of sparse graphs, such as insufficient information flow and fragmented representations. These results suggest that the advantages of NAS methods extend beyond modeling complex connectivity patterns and are also applicable to addressing the unique challenges posed by sparse graphs.

\begin{table*}[h]
\centering
\caption{Comparison of ABG-NAS with seven popular GNAS methods (GraphNAS, SNAG, SANE, Auto-HeG, DFG-NAS, DE, and PSO) across four datasets. The table reports the mean accuracy and macro-F1 score over 10 independent runs, with standard deviations included. ABG-NAS consistently achieves the highest performance across all datasets, demonstrating its robustness and effectiveness. Results marked with * are official results reported in their respective papers.}
\label{tab:table3_NAS_Comparison}
\scriptsize % 
\setlength{\tabcolsep}{4pt} 
\renewcommand{\arraystretch}{0.9} % 
\resizebox{\textwidth}{!}{
\begin{tabular}{cccccccccc}
\toprule
\multicolumn{2}{c}{\multirow{2}{*}{Methods}} & \multicolumn{2}{c}{Cora\cite{mccallum2000efficient}} & \multicolumn{2}{c}{PubMed \cite{sen2008collective}} & \multicolumn{2}{c}{CoraFull \cite{bojchevski2017deep}} & \multicolumn{2}{c}{Citeseer\cite{sen2008collective}} \\
\cmidrule(lr){3-4} \cmidrule(lr){5-6} \cmidrule(lr){7-8} \cmidrule(lr){9-10}
 & & \scriptsize Acc (\%) & \scriptsize Macro-F1 (\%) & \scriptsize Acc (\%) & \scriptsize Macro-F1 (\%) & \scriptsize Acc (\%) & \scriptsize Macro-F1 (\%) & \scriptsize Acc (\%) & \scriptsize Macro-F1 (\%) \\
\midrule
\multicolumn{2}{c}{GraphNAS* \cite{gao2021graph}} & 84.10±0.79 & - & 82.28±0.64 & - & - & - & 68.83±2.09 & - \\
\multicolumn{2}{c}{SNAG* \cite{zhao2020simplifying}} & 81.01±1.31 & - & 83.24±0.84 & - & - & - & 70.14±2.40 & - \\
\multicolumn{2}{c}{SANE* \cite{huan2021search}} & 84.25±1.82 & - & 87.82±0.57 & - & - & - & 74.33±1.54 & - \\
\multicolumn{2}{c}{Auto-HeG* \cite{zheng2023auto}} & 86.88±1.10 & - & 89.29±0.27 & - & - & - & 75.81±1.52 & - \\
\multicolumn{2}{c}{DE \cite{li2023population}} & 86.0±0.01 & 84.0±0.03 & 88.0±0.00 & 88.0±0.00 & 66.0±0.00 & 52.0±0.00 & 75.0±0.00 & 72.0±0.00 \\
\multicolumn{2}{c}{PSO \cite{zhong2024graph}} & 85.0±0.02 & 83.0±0.03 & 87.9±0.01 & 87.8±0.001 & 65.6±0.01 & 52.2±0.001 & 74.9±0.03 & 71.6±0.03 \\
\multicolumn{2}{c}{DFG-NAS\cite{zhang2022dfg}} & 86.6±0.03 & 84.6±0.05 & 88.9±0.01 & 88.7±0.01 & 70.4±0.03 & 60.9±0.05 & 77.2±0.05 & 73.4±0.06 \\
\midrule
\multicolumn{2}{c}{ABG-NAS (Our)} & \textbf{88.32±0.02} &  \textbf{87.52±0.05} & \textbf{89.88±0.04} & \textbf{89.67±0.02} & \textbf{72.14±0.06} & \textbf{63.03±0.07} & \textbf{77.44±0.02} & \textbf{73.50±0.01} \\\bottomrule
\end{tabular}}

\end{table*}

\subsection{Comparison with Diverse GNAS Methods}

In this experiment, we compared the performance of ABG-NAS with seven state-of-the-art GNAS methods across four datasets. These methods cover a range of frameworks, including RL-based (e.g., GraphNAS\cite{gao2021graph}), gradient-based (e.g., SANE\cite{huan2021search}), and EA-based approaches\cite{li2023population,zhong2024graph,zhang2022dfg}. As shown in Table \ref{tab:table3_NAS_Comparison}, ABG-NAS consistently achieves superior results on all datasets. For instance, on the Cora dataset, ABG-NAS achieves an accuracy of 88.32\%, outperforming the best baseline by 2.3\%.

Although ACC is a fundamental metric for evaluating model performance, to more fairly demonstrate the advantages of ABG-NAS, we further introduce Macro-F1 as a complementary evaluation metric to compare ABG-NAS with other widely-used EA-based NAS methods.Specifically, on the Cora dataset, our method achieved a Macro-F1 score of 87.52\%, significantly outperforming DFG-NAS (84.6\%), DE (84.0\%), and PSO (83.0\%). Similarly, on PubMed, ABG-NAS achieved 89.67\%, compared to 88.7\% for DFG-NAS, 88.0\% for DE, and 87.9\% for PSO. Even on the larger and more complex CoraFull dataset, ABG-NAS maintained stability with a Macro-F1 score of 63.03\%, outperforming methods that struggled on this dataset. Citeseer, with its high sparsity (\(D_{\text{avg}} = 2.84\)) and 217 isolated nodes (as shown in Table \ref{tab:table1_dataset_information}), poses a significant challenge for many NAS methods. Despite this, ABG-NAS achieves a Macro-F1 score of 73.50\%, outperforming DFG-NAS (73.4\%), DE (72.0\%), and PSO (71.6\%). This highlights ABG-NAS’s adaptability to sparse graphs, further showcasing its robustness across diverse graph structures.

\begin{figure}[H]
\centering
\includegraphics[width=\textwidth]{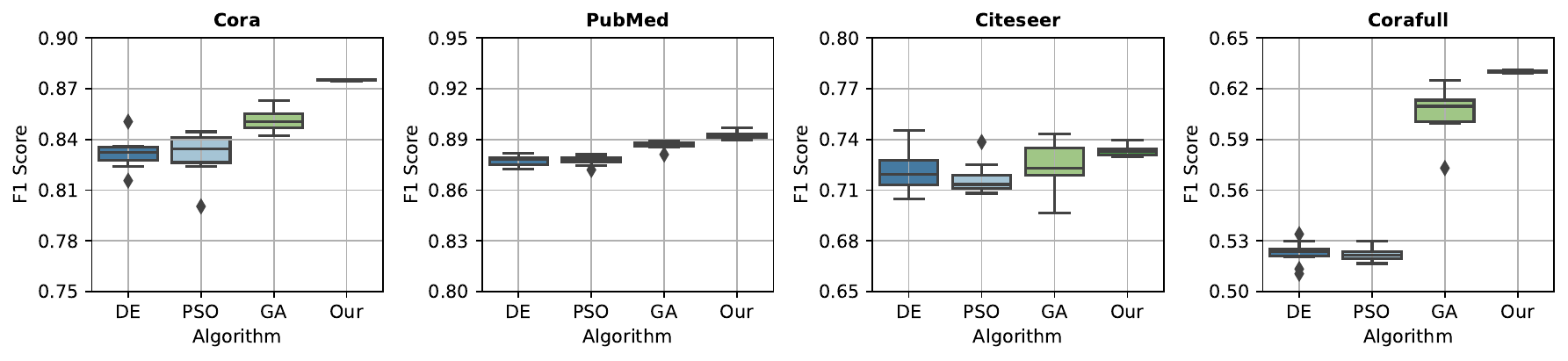}
\caption{Box plots showing the distribution of F1 scores over 10 runs for ABG-NAS and other EA-based NAS methods (PSO, DE, DFG-NAS) on four datasets. ABG-NAS consistently achieves higher median F1 scores with narrower interquartile ranges, indicating superior performance and stability across diverse graph datasets.}
\label{fig:comparison_f1_boxplot_1x4}
\end{figure}

\begin{figure}[H]
\centering
\includegraphics[width=\textwidth]{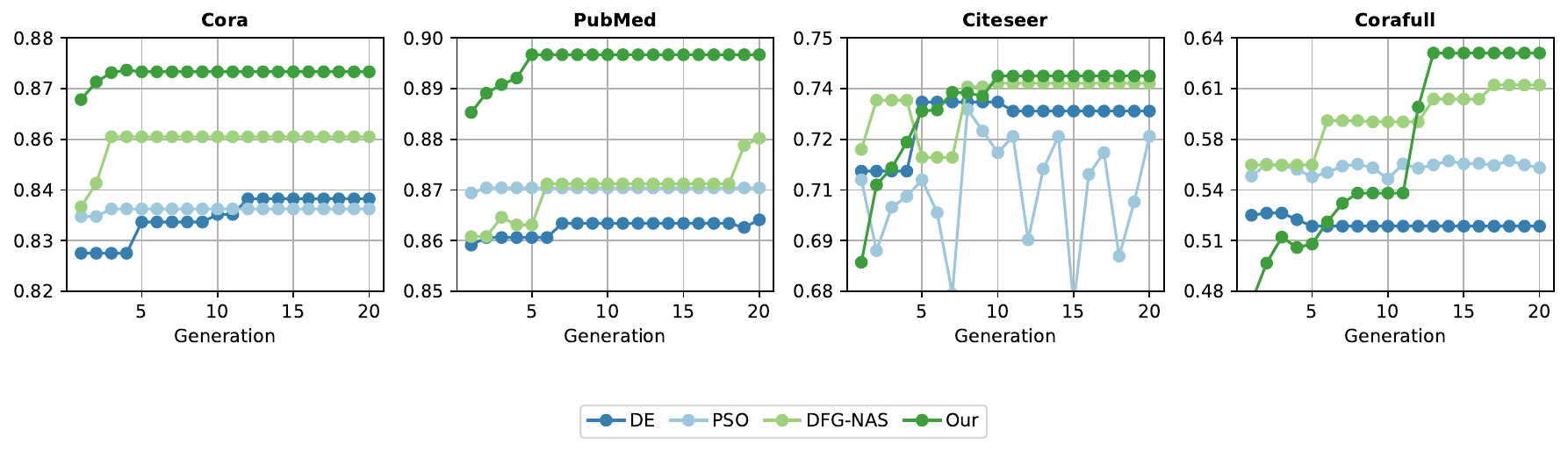}
\caption{Convergence curves of F1 scores over 20 generations for four datasets (Cora, PubMed, CiteSeer, and CoraFull) comparing ABG-NAS with other EA-based NAS methods (DFG-NAS, DE, and PSO). ABG-NAS demonstrates superior performance, achieving higher F1 scores on all datasets.}
\label{fig:comparison_f1_convergence_1x4}
\end{figure}

Figure \ref{fig:comparison_f1_boxplot_1x4} illustrates the distribution of F1 scores over 10 independent runs for each algorithm across four datasets. The results clearly demonstrate the superiority of our ABG-NAS method compared to other EA-based NAS methods (PSO, DE, DFG-NAS). ABG-NAS consistently achieves higher median F1 scores on all datasets, with notably narrower interquartile ranges, indicating less variation and greater stability. These results highlight the reliability and consistent effectiveness of ABG-NAS across diverse graph data.

Figure  \ref{fig:comparison_f1_convergence_1x4} visualizes the optimization process over 20 generations for a single run, which further highlight the effectiveness of ABG-NAS. On the Cora and PubMed datasets, our method achieves higher F1 scores from the very first generation and converges within the first few generations. On the more challenging Citeseer and CoraFull datasets, although our method starts with lower performance in the initial generation, it quickly surpasses the other methods and consistently converges to the highest F1 scores. Across all datasets, ABG-NAS demonstrates superior performance by consistently outperforming the other methods after convergence.

In summary, the results from both Table \ref{tab:table3_NAS_Comparison}  and Figures \ref{fig:comparison_f1_convergence_1x4} and \ref{fig:comparison_f1_boxplot_1x4} 
 underscore the advantages of ABG-NAS over state-of-the-art GA-based NAS methods (DFG-NAS), and other EA-based algorithms, i.e., DE, and PSO. Our method not only achieves higher accuracy and macro-F1 scores but also exhibits superior robustness and stability across diverse and complex graph structures. This makes ABG-NAS a more reliable choice for automated neural architecture search in GNNs, particularly for challenging real-world datasets.

\begin{table*}[h]
\centering
%\scriptsize %  
\setlength{\tabcolsep}{4pt} %  
\renewcommand{\arraystretch}{0.9} % 
\resizebox{\textwidth}{!}{
% \begin{tabular}{c ccc*{6}{>{\centering\arraybackslash}p{2.1cm}}}
\begin{tabular}{cccccccccc}
\toprule
\multicolumn{1}{c}{\multirow{2}{*}{Method}} & \multicolumn{3}{c}{Setting} & \multicolumn{2}{c}{Cora\cite{mccallum2000efficient}} & \multicolumn{2}{c}{PubMed \cite{sen2008collective}} & \multicolumn{2}{c}{CoraFull \cite{bojchevski2017deep}} \\
\cmidrule(lr){2-4} \cmidrule(lr){5-6} \cmidrule(lr){7-8} \cmidrule(lr){9-10}
 & CASS & AGOS & BGTM & \scriptsize Acc (\%) & \scriptsize Macro-F1 (\%) & \scriptsize Acc (\%) & \scriptsize Macro-F1 (\%) & \scriptsize Acc (\%) & \scriptsize Macro-F1 (\%) \\
\midrule
DFG-NAS\cite{zhang2022dfg} & \ding{55} & \ding{55} & \ding{55} & 86.96±0.06 & 85.17±0.07 & 88.87±0.02 & 88.66±0.02 & 70.4±0.09 & 60.60±0.01 \\
ABG-NAS-V1 & \ding{51} & \ding{55} & \ding{55} & 86.85±0.07 & 84.74±0.01 & 88.35±0.03 & 88.17±0.02 & 63.02±0.02 & 58.45±0.05 \\
ABG-NAS-V2 & \ding{55} & \ding{51} & \ding{55} & 88.21±0.05 & 86.46±0.07 & 89.24±0.01 & 89.09±0.01 & 70.50±0.01 & 61.16±0.02 \\
ABG-NAS-V3 & \ding{51} & \ding{51} & \ding{55} & \textbf{88.34±0.07} & 86.76±0.06 & 89.73±0.07 & 89.15±0.04 & 71.28±0.06 & 62.21±0.04 \\
\midrule
ABG-NAS & \ding{51} & \ding{51} & \ding{51} & 88.32±0.02 & \textbf{87.52±0.05} & \textbf{89.88±0.04} & \textbf{89.67±0.02} & \textbf{72.14±0.06} & \textbf{63.03±0.07} \\
\bottomrule
\end{tabular}}
\caption{Comparison of ablation study results on three datasets. Each NAS framework was executed 10 times, and the average accuracy and Macro-F1 scores with standard deviations are reported. The full ABG-NAS framework includes CASS, AGOS, and BGTM, while the variants exclude one or more components to assess their impact. The best results are highlighted in bold.}
\label{tab:table4_Ablation_Comparison}
\end{table*}

\begin{figure}[H]
\centering
\includegraphics[width=\textwidth]{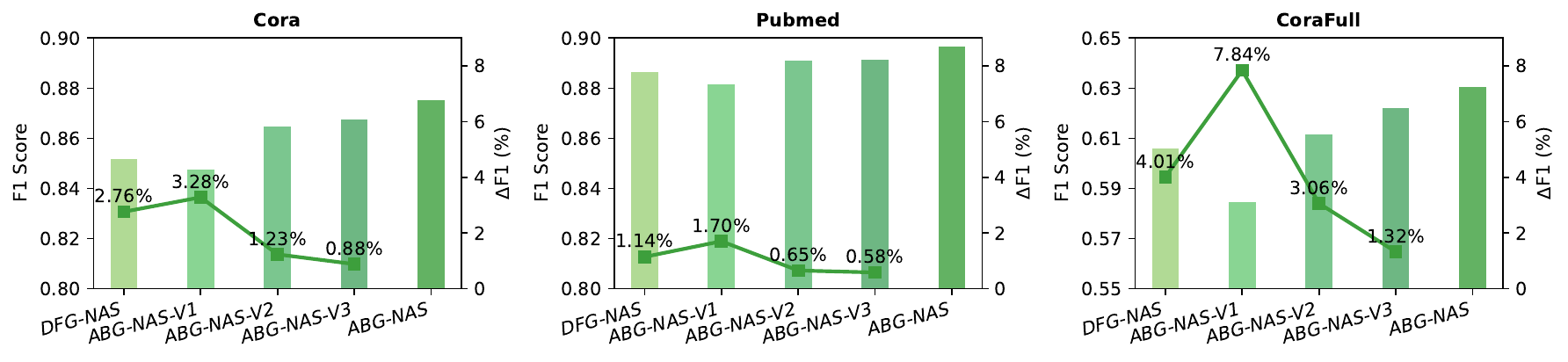}
\caption{Ablation study results showing F1 scores and percentage improvements for ABG-NAS variants across Cora, PubMed, and CoraFull datasets. Bar graphs illustrate F1 scores for each model variant, while line graphs depict the relative percentage improvements ($\Delta$F1) compared to the full ABG-NAS framework.}
\label{fig:ablation_scenarios}
\end{figure}
\subsection{Ablation Study} \label{subsec:ablationStudy}

Compared to the baseline DFG-NAS, ABG-NAS introduces three key components: CASS, AGOS, and BGTM. CASS increases the flexibility and adaptability of GNN architectures across different graph structures, AGOS enhances the search process by balancing exploration and exploitation through more advanced genetic operations, and BGTM fine-tunes the hyperparameters to improve performance. In this subsection, we conduct an ablation study to validate the effectiveness of these components. We compare the full ABG-NAS framework with its three variants: ABG-NAS-V1 (DFG-NAS + CASS), ABG-NAS-V2 (DFG-NAS + AGOS), and ABG-NAS-V3 (DFG-NAS + CASS + AGOS). The baseline DFG-NAS \cite{zhang2022dfg} is used as the reference point. The results are summarized in Table \ref{tab:table4_Ablation_Comparison}.

From Table \ref{tab:table4_Ablation_Comparison}, we observe that removing any component results in a consistent drop in Macro-F1 scores across all datasets, confirming the importance of each component in ABG-NAS. Notably, without the sophisticated AGOS search strategy, ABG-NAS-V1 (which only incorporates the expanded search space) performs worse than the baseline DFG-NAS across all datasets, indicating that CASS alone cannot drive improvements without an effective search strategy. In contrast, ABG-NAS-V2, which employs AGOS but lacks CASS, significantly outperforms DFG-NAS, highlighting AGOS's capability to enhance performance even in a simpler search space. When CASS is added to ABG-NAS-V2, forming ABG-NAS-V3, there is a slight improvement over ABG-NAS-V2, demonstrating the benefit of combining CASS with a more advanced search algorithm. Finally, incorporating BGTM into ABG-NAS-V3 produces the full ABG-NAS framework, which delivers the best results overall, validating the critical role of BGTM in hyperparameter fine-tuning for optimal performance.

We further visualize these results in Figure \ref{fig:ablation_scenarios}, which highlights the F1 scores and the percentage improvements achieved by ABG-NAS over DFG-NAS and the three variants. While the performance gap between variants is smaller on simpler datasets like Cora and PubMed, it becomes more pronounced on complex datasets such as CoraFull. For instance, the full ABG-NAS framework achieves a 4.01\% improvement in F1 score over DFG-NAS, a 7.84\% improvement over ABG-NAS-V1, a 3.06\% improvement over ABG-NAS-V2, and a 1.32\% improvement over ABG-NAS-V3 on CoraFull. These results underscore that ABG-NAS, as a comprehensive and robust framework, is capable of addressing diverse graph structures and tasks, delivering superior performance across various datasets.

\section{Conclusion}
\label{sec:conclusion}

In this paper, we introduce ABG-NAS, a novel framework designed to efficiently optimize neural architecture search for improved graph representation learning. ABG-NAS seamlessly integrates three components: Comprehensive Architecture Search Space (CASS), Adaptive Genetic Optimization Strategy (AGOS), and Bayesian-Guided Tuning Module (BGTM). These components synergistically explore and exploit a diverse range of GNN architectures while jointly optimizing hyperparameters, effectively adapting to graph datasets with varying sparsity. Our framework is both model-agnostic and scalable with respect to graph size and architectural complexity, enabling it to support downstream graph learning tasks across a variety of real-world datasets. Empirical evaluations on various benchmark datasets demonstrate that ABG-NAS consistently outperforms both manually designed GNN and existing state-of-the-art NAS methods. The results of our ablation study further validate the effectiveness of the composite design of our framework.

In the future, we aim to further generalize the search space to better handle more complex and heterogeneous graph structures, while ensuring continued efficiency and scalability across diverse graph sizes and model complexities. 
We also plan to conduct a comprehensive evaluation of ABG-NAS's computational efficiency—such as GPU time and convergence speed—under unified benchmarking settings to strengthen the analysis of its practical viability. Furthermore, we plan to develop an explainability layer to streamline the interpretability of learned representations, further enhancing the framework's usability in practical applications.

\section*{Acknowledgement}

This work was supported in part by the Australian Research Council under Discovery Project Grant No. DP230100716.

\bibliographystyle{elsarticle-num} 
\bibliography{Manuscript_final_submit}

\end{document}